# A Flexible FBG-Based Contact Force Sensor for Robotic Gripping Systems*

Wenjie Lai [1,2], *IEEE Member*, Huu Duoc Nguyen [1,2], Jiajun Liu[1], *IEEE Member*,
Xingyu Chen [1,2], *IEEE Member,* Soo Jay Phee [1,2], *IEEE Senior Member*

*Abstract*— Soft robotic grippers demonstrate great potential for gently and safely handling objects; however, their full potential for executing precise and secure grasping has been limited by the lack of integrated sensors, leading to problems such as slippage and excessive force exertion. To address this challenge, we present a small and highly sensitive Fiber Bragg Grating-based force sensor designed for accurate contact force measurement. The flexible force sensor comprises a 3D-printed TPU casing with a small bump and uvula structure, a dual FBG array, and a protective tube. A series of tests have been conducted to evaluate the effectiveness of the proposed force sensor, including force calibration, repeatability test, hysteresis study, force measurement comparison, and temperature calibration and compensation tests. The results demonstrated good repeatability, with a force measurement range of 4.69 N, a high sensitivity of approximately 1169.04 pm/N, a root mean square error (RMSE) of 0.12 N, and a maximum hysteresis of 4.83%. When compared to a commercial load cell, the sensor showed a percentage error of 2.56% and an RMSE of 0.14 N. Besides, the proposed sensor validated its temperature compensation effectiveness, with a force RMSE of 0.01 N over a temperature change of 11 °C. The sensor was integrated with a soft grow-and-twine gripper to monitor interaction forces between different objects and the robotic gripper. Closed-loop force control was applied during automated pick-and-place tasks and significantly improved gripping stability, as demonstrated in tests. This force sensor can be used across manufacturing, agriculture, healthcare (like prosthetic hands), logistics, and packaging, to provide situation awareness and higher operational efficiency.

## I. Introduction

BIO-INSPIRED soft robots have recently attracted increasing attention, due to their unique features such as safety, high adaptability, ease of fabrication, and relatively low cost [1], [2]. However, soft robots exhibit inherent instability arising from their high dexterity and mechanical compliance, which poses significant challenges in control complexity, reduced operational speed, and decreased precision compared with rigid robots [1]. For soft grippers, it is important to prevent slippage to maintain grasping efficiency. Also, successfully grasping an object on the first attempt without excessive force is essential, as it minimizes energy consumption, extends the lifespan of both the gripper and the object, and avoids damage [3]. Therefore, integrating sensors with soft gripper to monitor contact status is crucial to prevent slippage and damage during robot manipulations. Nevertheless, integrating sensors into soft robots to achieve accurate force perception remains challenging due to constraints in size, high sensitivity and resolution, reliable repeatability, material compliance, and the need to decouple disturbances from temperature fluctuations, vibrations, and shape changes [4], [5]. To address this challenge, we developed a small, highly responsive and sensitive soft force sensor with temperature compensation feature that can be easily attached to grippers to realize the precise closed-loop force control during robotic manipulation tasks.

To monitor contact force and detect slippage, recent research has investigated sensing material [6], [7] and various sensing technologies, including resistance-based sensors [5], optoelectronic sensors [8], pressure sensors [9], MEMS (Micro-Electro-Mechanical Systems) [10], and optical sensors [3], [11], [12]. Among these sensing methods, optical sensing appears as an effective and compelling approach for soft grippers, because of its high sensitivity, rapid response, electrical passivity, and ease of integration [13]. In 2020, James *et. al.* embedded optical tactile sensors in a three-fingered hand to perform slip detection for eleven different objects with masses from 100 to 300 grams [3]. In 2021, Jamil *et. al* used hybrid intensity-based optical fiber sensors to measure both bending and contact force for a soft pneumatic gripper [11]. However, their observable hysteresis issue may result in measurement inaccuracies. Fiber Bragg Grating (FBG) sensors have been recently explored in soft grippers due to their structural simplicity and minimal influence from phase discontinuity or intensity fluctuations, which affect other types of optical fiber sensors. In 2024, Hou *et. al* directly sticked an FBG sensor into each flexible fingertip to monitor the contact status, offering slip prediction about 0.7 seconds before the critical slip point [14]. In 2023, Lyu *et. al* attached ultra-short FBG (grating length: 1 mm) with a fin-ray finger to realize static-dynamic simultaneous detection of incipient slip [15]. In those works, optical fibers are exposed to the environment without proper protection, which may shorten the sensor lifespan and increase the risk of breakage. In 2020, Yang *et. al* embedded FBG sensors to a silicone layer in sinusoidal configuration to realize slippage detection in a soft robotic gripper, tested with spherical objects weighing 15-41 grams [16]. Most FBG-based sensors are embedded within subtrat and it is difficult to decouple effects from bending and force. Besides, current implementations reveal certain limitations, underscoring the need for smaller

*This research is supported by grant (021990-00011) from the National Research Foundation, Prime Minister's Office, Singapore under its Campus of Research Excellence and Technological Enterprise (CREATE) programme.

Affiliation:
1. School of Mechanical and Aerospace Engineering, Nanyang Technological University, Singapore.
2. Singapore-HUJ Alliance for Research and Enterprise (SHARE), The Smart Grippers for Soft Robotics (SGSR) Programme, Campus for Research Excellence and Technological Enterprise (CREATE), Singapore.

Corresponding author: W. Lai (wjlai@ntu.edu.sg ).

and protective designs, higher sensitivity and improved tolerance to unwanted disturbances for contact force detection.

In this work, we introduced a compact and highly sensitive FBG-based force sensor designed for precise contact force measurement in robotic gripping systems. Unlike conventional soft sensor approaches that bond the FBG sensor to a thin-layer substrate and measure force through fiber stretching along the longitudinal direction, our design suspends the fiber within a chamber, allowing direct measurement of transverse forces on the FBG sensor. This new configuration enhances sensitivity and measurement accuracy. Our proposed soft force sensor consists of a 3D-printed Thermoplastic polyurethane (TPU) casing with a small bump and uvula structure, a dual FBG array, and a protective tube. Through tests, the proposed sensor shows good repeatability, with a force measurement range of 4.69 N, a high sensitivity of approximately 1169.04 pm/N, a root mean square error (RMSE) of 0.12 N, and a maximum hysteresis of 4.83%. When compared to a commercial load cell, the sensor showed a percentage error of 2.56% and an RMSE of 0.14 N. Besides, the sensor validated its temperature compensation effectiveness, with an RMSE of 0.01 N over a temperature change of 11 °C. The force sensor was subsequently integrated with our previously developed soft grow-and-twine gripper [18], which adapts to objects of varying weights, shapes, and sizes, allowing the monitoring of interaction forces between the object and the gripper. Closed-loop force control, facilitated by the sensor, was applied during automated pick-and-place tasks involving three objects of various materials and weights. Comparative tests demonstrated a significantly improvement in gripping stability.

The rest of this article is organized as follows. *Section II* presents the development of the proposed force sensor, including design and fabrication. *Section III* elaborates on the working principles, covering force-fitting and temperature compensation. *Section IV* provides the evaluation tests and results for force calibration, repeatability, hysteresis analysis, force measurement comparison, as well as temperature calibration and compensation test. *Section V* illustrates the integration of the sensor with a soft gripper and a robotic arm, along with an evaluation test of grasping stability during automated pick-and-place demonstrations, comparing performance with and without the closed-loop force control. Finally, *Section VI* concludes the article.

## II. DESIGN AND FABRICATION

### A. Model Design of the Force Sensor

The FBG-based force sensor is designed to measure the contact force $N$ applied to a small bump on the surface of the casing (*Fig. 1*). This compact device has dimensions of 12 mm x 12 mm x 4 mm, consisting of a 3D-printed flexible casing made from TPU, an optical fiber with a dual FBG array (Technica; grating length: 1 mm; central wavelengths (CW): 1540/1550±0.5 nm; full width at half maximum (FWHM): 0.6-0.7 nm; Reflectivity: > 30%; FBG profile: Apodized; Recoating: Polyimide), and a protective tube *Hytrel®* (inner diameter (ID): 0.4 mm; outer diameter (OD): 0.9 mm). The optical fiber is fully enclosed within the casing and protective tube, shielding them from exposure to the surrounding environment and thereby improving the robustness of the sensor.

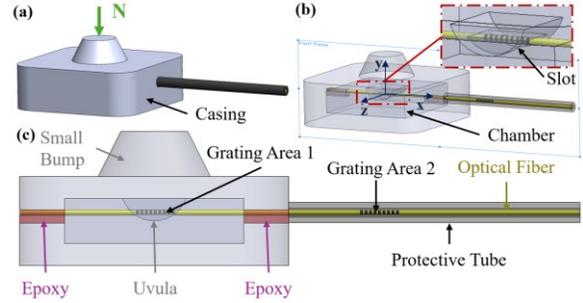

**Fig. 1.** (a) 3D model of the force sensor. (b) Transparent view highlighting the slot feature. (c) Details of the force sensor shown in the cross-section view.

Inside the casing, a chamber (8 mm x 8 mm x 2 mm) provides space for the uvula to move downward when the bump is pressed. The gap between the uvula and the bottom surface inside the chamber is 1 mm. The small bump is designed to transmit and focus the applied force, ensuring its efficient transmission to the uvula, which causes deformation in the FBG sensor (*FBG1 at Grating Area 1*). The slender layer between the bump and the uvula is 1 mm thick. The optical fiber is suspended beneath the slot (width: 1.5 mm; depth: 0.5 mm) aligned with the uvula, with both ends secured using *UHU Epoxy Ultra Strong*. The slot constraints the downward movement of the optical fiber while providing enough clearance to minimize interference from small forces along the x- and z-axes (*Fig. 1b*). One FBG sensor (*FBG1 at Grating Area 1*) is positioned directly beneath the uvula to detect strain changes from the applied contact force, and the resulting wavelength shift $\Delta\lambda_B$ of the FBG sensor is automatically recorded by an interrogator (*Luna HYPERION si255*). A second FBG sensor (*FBG2 at Grating Area 2*) is housed in the protective tube, remaining strain-free to serve as a temperature sensor, compensating for thermal effects. The FBG2 sensor is 10 mm away from the FBG1 sensor.

### B. Fabrication of the Force sensor

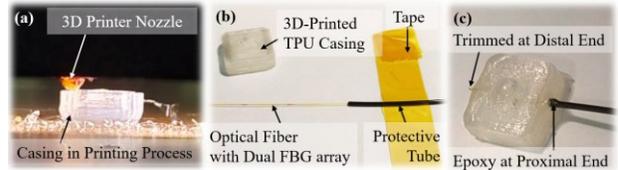

**Fig. 2.** Fabrication of the force sensor: (a) 3D printing of the casing. (b) Preparation of the components. (c) Prototype of the assembled force sensor.

The casing was printed using soft material (*NinjaFlex Water Semi-Transparent TPU*) by an FDM 3D printer (*Original Prusa XL*) (*Fig. 2a*). The printing parameters were configured as follows: nozzle diameter: 0.4 mm; nozzle temperature: 235 °C; printing bed temperature: 60 °C; layer height: 0.2 mm; printing width: 0.6 mm.

*Fig. 2b* shows all the components required to assemble the force sensor. The section of the optical fiber containing the sensor FBG1 was inserted into the casing. Epoxy was applied to both ends of the casing using a 0.4-mm diameter needle. To maintain pretension in the optical fiber during the curing process, tape (*Polyimide Adhesive Tape*) was applied to both ends of the fiber. The remaining portion of the fiber was then inserted into a protective tube that extended along the entire pigtail to one end of the casing, ensuring no bare fiber was exposed. Additional epoxy was applied to the proximal end of the casing to secure the assembly. At the distal end of the casing, the excess fiber was trimmed at an inclined angle to keep the reflectivity of the FBG sensors. A prototype picture to show the final assembled force sensor (*Fig. 2*c).

## III. WORKING PRINCIPLES

FBGs are microstructures photo-inscribed in the core of a single-mode optical fiber, reflecting specific light spectra in response to strain and temperature changes, which cause physical deformation of the fiber. The central wavelength of the FBG reflection band is called the Bragg wavelength $\lambda_B$, as in [20]:

$$\lambda_B = 2\, n_{eff}\Lambda, \qquad (1)$$

where $n_{eff}$ is the effective refractive index of the optical fiber and $\Lambda$ is the pitch of refractive index modulation. $\Lambda$ depends on the temperature and strain experienced by the sensor in the grating area.

The wavelength shifts due to changes in strain and temperature, as in [21]:

$$\Delta\lambda_B = K_T \Delta T + K_\varepsilon \Delta\varepsilon, \qquad (2)$$

where $K_T$, $\Delta T$, $K_\varepsilon$, and $\Delta\varepsilon$ are the temperature sensitivity, temperature change, strain sensitivity, and strain change in the grating area along the fiber, respectively.

For this dual FBG array, the wavelength shift $\Delta\lambda_{B,1}$ of the sensor FBG1 inside the chamber results from both $\Delta\lambda_{B,T1}$ and $\Delta\lambda_{B,\varepsilon1}$, reflecting changes in temperature and strain caused by the applied contact force $N$. While the strain-free FBG2 within the protective tube experiences wavelength shift $\Delta\lambda_{B,T2}$, solely related to temperature changes. Therefore, we have:

$$\Delta\lambda_{B,1} = K_{T_1}\Delta T_1 + K_{\varepsilon_1}\Delta\varepsilon_1 = \Delta\lambda_{B,T_1} + \Delta\lambda_{B,\varepsilon_1}, \qquad (3)$$

$$\Delta\lambda_{B,2} = K_{T_2}\Delta T_2 = \Delta\lambda_{B,T_2}, \qquad (4)$$

where $K_{T1}$, $\Delta T_1$, $K_{\varepsilon 1}$, $\Delta\varepsilon_1$, $K_{T2}$, and $\Delta T_2$, are the temperature sensitivity, temperature change, strain sensitivity, strain change on the sensor FBG1, and the temperature sensitivity, temperature change on the sensor FBG2, respectively. The ratio between temperature sensitivities of sensors FBG1 and FBG2 is $r = K_{T1}/K_{T2}$. The temperature sensitivities and ratio can be obtained through experiment as shown in *Section IV*.

Since both sensors are close to each other, the temperature change experienced by these two sensors is assumed the same, $\Delta T_1 = \Delta T_2$. Thus, we have:

$$\Delta\lambda_{B,1} = r\Delta\lambda_{B,T_2} + \Delta\lambda_{B,\varepsilon_1} = r\Delta\lambda_{B,T_2} + K_{\varepsilon_1}\Delta\varepsilon_1, \qquad (5)$$

For a cantilever beam fixed at both ends and loaded at center, large deformation causes a nonlinear response due to combined bending and stretching, making the displacement significantly deviate from simple linear predictions [22]. In this scenario, a second-order polynomial equation is used to model the nonlinear relationship between the applied force $N$ and stain change $\Delta\varepsilon_1$, where $a$, $b$, $c$ are constant:

$$\Delta\varepsilon_1 = aN^2 + bN + c. \qquad (6)$$

Equation (6) can be further derived as:

$$AN^2 + BN + C - (\Delta\lambda_{B,1} - r\Delta\lambda_{B,T_2}) = 0, \qquad (7)$$

where $A$, $B$, $C$ are constant and can be obtained through experiment at lab temperature when $\Delta\lambda_{B,T2} = 0$, as shown in *Section IV*.

The contact force $N$ can be then calculated as:

$$N = \frac{-B + \sqrt{(B^2 - 4A[C - (\Delta\lambda_{B,1} - r\Delta\lambda_{B,T_2})]}}{2A}, \qquad (8)$$

As a result, temperature influence on the sensor FBG1 can be effectively offset from the measurement of the contact force, which is proved through experiment in *Section IV*.

## IV. EVALUATION EXPERIMENTS AND RESULTS

### A. Experimental Setups

A series of tests were conducted to evaluate the performance of the assembled force sensor, including force calibration and repeatability tests, a hysteresis study, a force measurement comparison, as well as temperature calibration and compensation tests.

In the test platform related to force measurement, the force sensor was placed on top of a load cell (*Futek, LTH 300*) to measure the ground truth value of the transmitted contact force $N$ (*Fig. 4a* and *Fig. 4b*). A linear actuator, mounted on a fixture above the force sensor, was controlled by a microcomputer (*Raspberry Pi 3B*) to perform cyclical movements. It moved downward to apply a load and then upward to unload the proposed force sensor. As the tip of the linear actuator touched the small bump of the sensing module, the load cell recorded the applied force $N$ from the linear actuator, while data was collected through a data acquisition board (*Arduino Uno*). Simultaneously, the interrogator captured the corresponding wavelength shift $\Delta\lambda_B$ of the sensor FBG1. Four test runs were conducted to determine the relationship between the wavelength shift and the applied force. After obtaining the fitted calibration equations, an additional three test runs were performed to compare the force measurement accuracy with the ground truth values provided by the load cell.

To take the temperature effect into account, temperature calibration and compensation tests were carried out on the force sensor (*Fig. 4c* to *Fig. 4e*). In the temperature calibration setup, the force sensor was immersed in the water in a lab glass that sat on a hot plate heater (*IKA HCT basic*). The water was heated up by the heater, with the temperature measured using a digital thermometer embedded in the heater. During the heating process, the corresponding wavelength shifts of both FBG1 and FBG2 sensors were reflected by the interrogator. Based on these data, the temperature sensitivities, $K_{T1}$ and $K_{T2}$, as well as their sensitivity ratio $r$, were then determined. Subsequently, a temperature compensation test was conducted to validate the compensation model in *Subsection III.A*. For this test, tape was used to secure the force sensor to the inner surface of the lab glass, ensuring consistent contact force. The water was then gradually heated up by the heater. Force measurements with and without temperature compensation were calculated and compared to assess the effectiveness of the compensation approach.

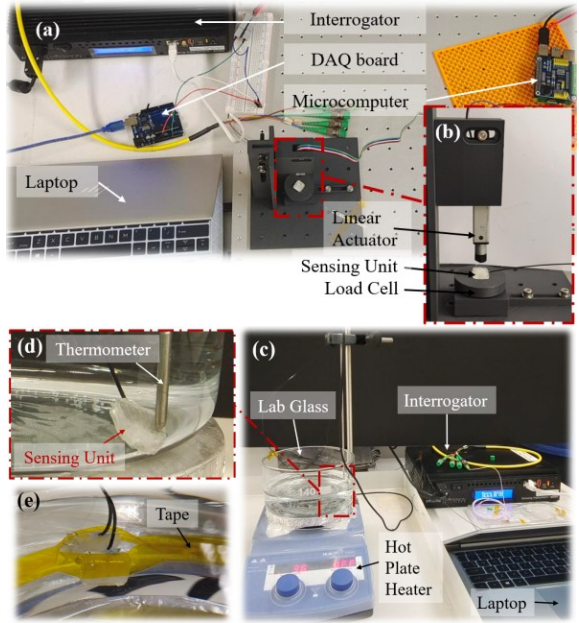

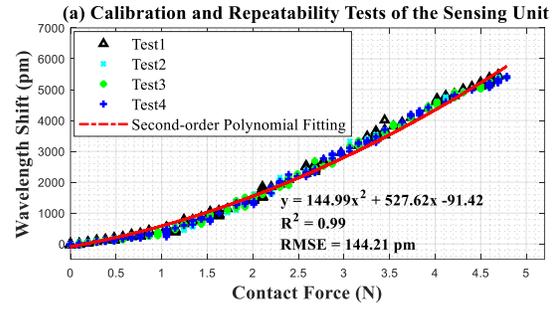

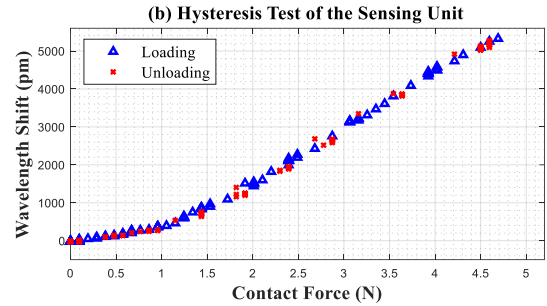

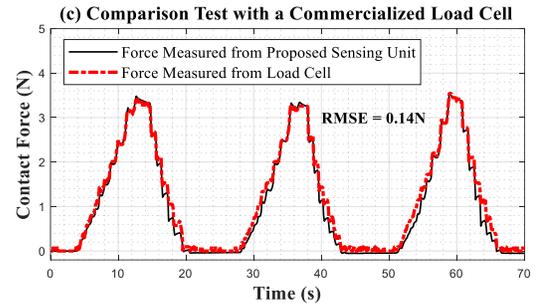

**Fig. 3.** (a) Top view of the experimental setup used for force calibration, repeatability, hysteresis, and comparison tests of the FBG1 sensor. (b) Zoomed-in side view of the force measurement platform. (c) Front view of the experimental setup for temperature calibration and compensation tests involving the FBG1 and FBG2 sensors. (d) Close side view of the force sensor submerged in a water bath. (e) Detailed view showing tape used to ensure consistent contact force on the force sensor during the temperature compensation test.

The results of force calibration, repeatability tests, hysteresis study, force comparison study, temperature calibration, and compensation tests can be found in the following subsection.

*B. Test Results*

Four cycles of loading and unloading tests were conducted using the linear actuator to evaluate the repeatability of the assembled force sensor. Data from all four runs showed close overlap, which indicates the good repeatability of the force sensor (*Fig. 5a*). A second-order polynomial fit of the data yield the calibration equation: $y = 144.99x^2 + 527.62x - 91.42$, with an $R^2$ value of 0.99 and a RMSE of 144.21 pm. Thus, *A, B, C* in equation (7) were obtained. The applied contact force varied from 0 N to 4.69 N, resulting in a wavelength shift between 0 pm and 5482.78 pm on the proposed force sensor. The sensor demonstrated high sensitivity at approximately 1169.04 pm/N, with a measurement error of around 0.12 N. This corresponds to a force measurement percentage error of approximately 2.56%. The interrogator used for data acquisition provides a resolution of 1 pm and a noise level of around 3 pm. Correspondingly, the force resolution of the force sensor is approximately 0.86 mN, while the signal-to-noise ratio is around 2.57 mN.

A force hysteresis analysis was then conducted, where maximum hysteresis 4.83 % occurred at 2.68 N (*Fig. 5b*). This hysteresis behavior can be attributed to the inelastic properties of the 3D-printed material [23] and the epoxy used during fabrication. To minimize hysteresis, we may consider optimizing material selection (e.g., using low-creep polymers and high-performance adhesives), refining fabrication methods, or implementing software-based hysteresis compensation algorithms.

**Fig. 4.** Results for the force sensor: (a) Repeatability tests and calibration test. (b) Hysteresis study of the force sensor. (c) Comparison test between the force sensor and a commercialized load cell.

The force calibration equation was used to convert the wavelength shift into corresponding contact force values. To assess the performance of the force sensor, three additional cyclic tests were conducted alongside a commercial load cell, resulting in an RMSE of 0.14 N (*Fig. 5c*). This indicates that the calibration equation is effective and shows the reliability of the force sensor for contact force measurement. *Section IV* presents an integration demonstration that illustrates how this force sensor facilitates closed-loop control in pick-and-place tasks.

During the temperature test, the water was heated from room temperature (25 °C) to 45 °C, with the entire force sensor, including the FBG1 and FBG2 sensors, freely submerged in the water within the lab glass. This range of 25 °C to 45 °C captures the usual variations in environmental temperature. Using the linear fitting, the temperature sensitivities for both sensors were obtained, where $K_{T1} = 24.29\ pm/°C$ and $K_{T2} = 10.31\ pm/°C$, with $R^2$ values of 0.99 and 1.00, respectively (*Fig. 6a*). The force sensor has good linearity between temperature changes and wavelength shifts.

After determining the temperature sensitivity ratio $r = 2.356$ (24.29 / 10.31), this value was applied for FBG1 in the temperature compensation test. Initially, the force sensor was placed freely in the water at a temperature of 34 °C. Tape was subsequently used to secure the force sensor to the inner surface of the glass. During the application of the tape, the force sensor experienced a sudden pressure from a finger around the *10th second*. Upon removal of the finger, a small residual force was exerted on the unit due to the tension of the tape. The residual force gradually reduced and stabilized at 0.1 N at the *32nd second*. At the meantime, the water was gradually heated to 45 °C. As illustrated in the results (right y-axis in *Fig. 6b*), the force sensor underwent a temperature change of 11 °C during this temperature compensation test, as indicated by the readings from the strain-free sensor FBG2. Without accounting for temperature effects, the measured force would have increased from 0.1 N to 0.5 N for sensor FBG1 (left y-axis in *Fig. 6b*), leading to a maximum measurement error of 0.4 N. With temperature compensation using equation (8) derived in *Section III*, the force measured by the sensor FBG1 fluctuated around the saturated value of 0.1 N from the *32nd second* onward. The RMSE with temperature compensation was 0.01 N, whereas the RMSE without temperature compensation was 0.23 N. This outcome validates the effectiveness of the temperature compensation feature of the force sensor.

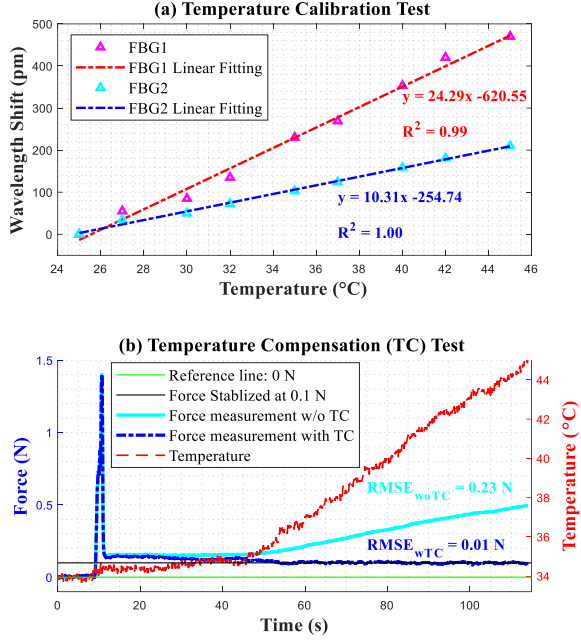

**Fig. 5.** Results for the force sensor: (a) Temperature calibration test. (b) Temperature compensation test.

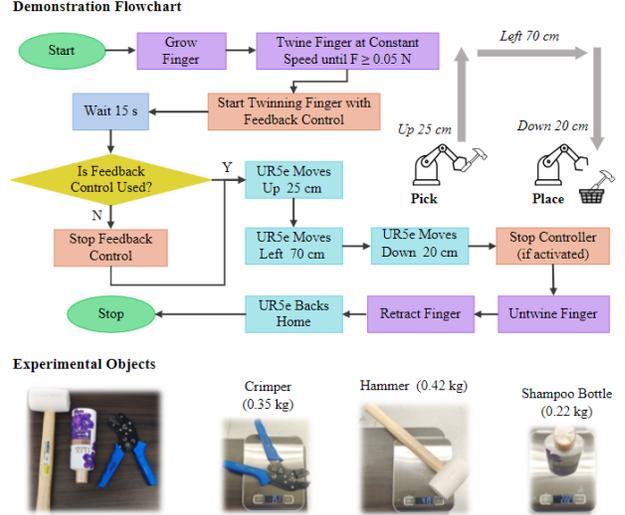

**Fig. 6.** Pick-and-Place Task Flowchart and Experimental Objects. The pick-and-place task is set up to compare the grasping performance with and without feedback control. In the latter case, the controller is deactivated before the robotic arm starts moving toward the placement location. A 15 s stabilization period is included for the controller to attain a firm grip, though a conditional force-based check could be implemented in future work.

## V. Evaluation of Grasping Stability in a Pick-and-Place Demonstration

### A. System Architecture and Integration

The force sensor was integrated into an automated robotic system along with a previously designed grow-and-twine gripper [18] and an UR5e robotic arm to perform automated pick-and-place tasks (*Fig. 7a*). The adaptive gripper utilizes the force sensor feedback to enhance grasping stability through closed-loop force controllers. The system was developed using the Robotics Operating System 2 framework (ROS2, Humble), which features a modular design for better scalability and maintainability. The system hardware consisted of two Linux processors (*Ubuntu 22.04*) communicating via Ethernet (*Fig. 7b*). One processor, equipped with real-time capabilities (*PREEMPT_RT patch*), ensured stable and low-latency control of the UR5e manipulator [24], [25]. The second processor, featuring an Nvidia driver (*GeForce RTX 3060, CUDA 12.1*) for future computer vision integration, interfaced with the gripper via its motors (*Dynamixel MX106*) and the sensor via an interrogator (*Luna HYPERION si255*). The system software comprised various ROS2 modular nodes, including a "Scheduler Node" for managing task sequences, a "UR5e Node" using MoveIt2 for manipulator control, and a closed-loop control system comprising a "Feedback Control Node," a "Gripper Node," and a "Sensor Node" to ensure stable grasping (*Fig. 7c*).

### B. Automated Pick-and-Place Task Setup

An automated pick-and-place task was designed to evaluate how the proposed force sensor improves grasping stability (*Fig. 8*). Three different objects of various weights, materials and shapes were chosen, including a crimper with rubber handler (351g), a wooden hammer (419g), and a plastic shampoo bottle (222g). A simple PID controller was implemented for the gripper, utilizing force feedback to regulate the contact force by adjusting the twining curvature of the gripper. The controller gains were set to $K_p = 20$, $K_i = 0.05$, and $K_d = 0$, experimentally determined to achieve 90% of the desired contact force within 5 s.

In the pick-and-place task, the UR5e robotic arm was first initialized at its home position. With a 14.0 cm growth extension, the soft finger then grasped each object, utilizing a 15 s grasping period to ensure a firm grip and force convergence. Setpoints for the contact force were set at 1.6 N for the hammer, 0.8 N for the shampoo bottle, and 0.6 N for the crimper (*Fig. 9*). These thresholds are determined based on a force balance between the weight and friction, while also considering the number of contact points. After that, the robot moves up to transfer the object to the collection site.

During the movement, disturbances (e.g. gravitational and inertial forces, vibrations) would be introduced to the gripper and its controller. Two scenarios were compared to highlight how the force sensor and closed-loop control improve grasping stability: 1) deactivating the force controller before the arm movement; 2) maintaining the closed-loop control throughout the pick-and-place task. The maximum joint velocity and acceleration of the UR5e manipulator during transfer were all limited to 45 deg/s and 15 deg/s$^2$, respectively.

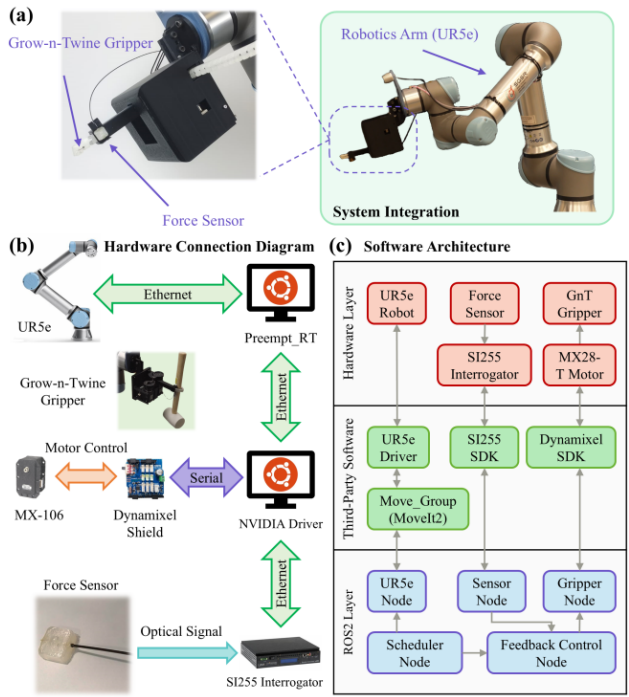

**Fig. 7.** System Architecture. (a) The automated robotic system, integrating a soft gripper, a force sensor, and a UR5e robotic arm. The force sensor, mounted at the soft finger's sleeve, measures contact force as the finger twines around objects. (b) Hardware architecture, featuring modular design employing two processors, enables independent operation and specialized functionalities. (c) Software architecture based on ROS2, enabling inter-node communication, modular functionality, and interaction with third-party SDKs.

### C. Enhanced Grasping Stability with Feedback Control

The sensor monitors the gripping status between the gripper and the item. When the item starts to slip, the force reading gradually decreases. To prevent slippage, we implement a closed-loop force control system that adjusts the grip in real-time, ensuring a secure grasp throughout pick-and-place operations.

The lack of force feedback leads to instability in grasping, causing all three objects to slip during the pick-and-place task (*Fig. 9-10*), due to the force imbalance throughout the movements. When the item is slipping, the force reading is gradually dropping (*Fig. 9(1)-(3)*). The crimper was dropped during the robot's upward motion (at time ~10s), while the shampoo bottle slipped during the horizontal transfer period (at time ~13s). The hammer, gripped with a higher initial force, slipped and fell as the arm approached the collection site (at time ~17s). In each case, the force readings rapidly decreased to zero as the objects slipped, indicating a loss of contact between the soft finger and the objects due to disturbances. Conversely, incorporating feedback control significantly improved grasping stability, as the PID controller allowed the gripper to retain all three objects throughout the task (*Fig. 9-10*). The contact force oscillates around its setpoints, demonstrating the controller's active effort to compensate for the unexpected disturbances. For example, when lifting the crimper, the controller successfully recovered the contact force to its setpoint after a momentary drop to 0.43 N (28%, *Fig. 9*). The measured force only fell to zero upon the intentional release of the objects at the destination.

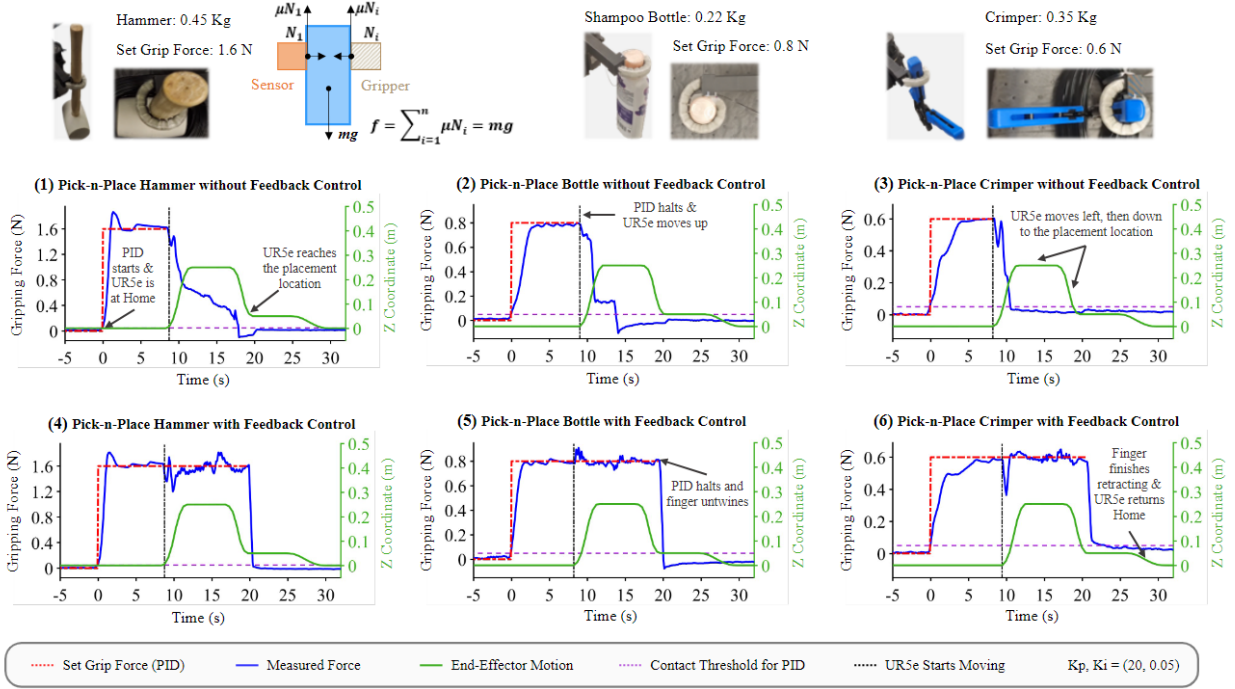

**Fig. 8.** Closer Look at Grasping Stability with Feedback Control. The desired contact force is set based on object weight and material properties, considering that the finger makes multiple contact points with the object, establishing a net friction force to counter the object's weight. Each chart presents the interplay of the PID controller (red), measured contact force (blue), and robot arm motion (green). Without feedback control, the contact force decreases when the arm starts moving (marked by black). With feedback control, the PID controller actively regulates contact force around the setpoint, allowing object release only at the designated placement location (green curve completed descending).

While implementing PID controller strengthens the grasping stability, there remains room for improvement. Though maintained throughout the pick-and-place task, the contact force fluctuated largely around the setpoint. The largest fluctuations reached 25% for the hammer, 12% for the shampoo bottle, and 28% for the crimper, highlighting the limitations of the controller in compensating for the disturbances (*Fig. 9*). An adaptive gain scheduling strategy, dynamically adjusting the gains based on the arm's motion (e.g., position, velocity, and acceleration) [26], could enhance force regulation, especially in high-speed applications. Additionally, the controller performance varied across objects with different shapes and materials, even with fixed gains. For instance, an overshoot of approximately 12% occurred when grasping the hammer but not for the other two objects (*Fig. 9*). Integrating vision-based object recognition, e.g., YOLO models, could enable the system to select appropriate control gains based on the object characteristics [27]. Such an ability would also facilitate the determination of suitable contact force setpoint, recognizing that smoother surfaces (e.g., PP or PE of the shampoo bottle) may require higher contact force compared to rougher surfaces (e.g., TPR or PVC of the crimper handler), even when lighter. In addition, further enhancements could involve optimizing the finger length to increase contact area and contact force, potentially using vision-based object detection to estimate object dimensions [28]. Furthermore, the soft material used in the finger introduced nonlinearities, necessitating advanced control techniques beyond the linear-system-favored PID controller. These may include gain scheduling or methods like optimization-based, model-based, or sliding mode control, and machine learning [29], [30]. Besides, relocating the force sensor along the finger instead of at the sleeve could provide more precise feedback on the distributed contact force.

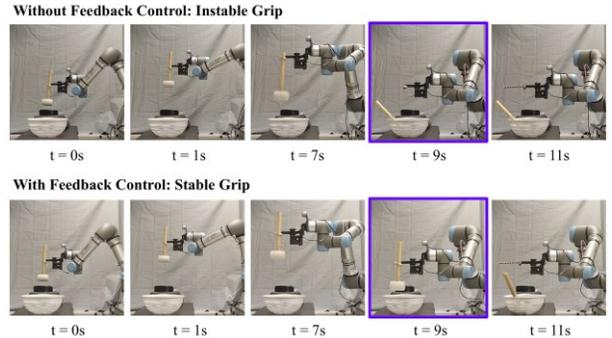

**Fig. 9.** Force Feedback Control Enables Stable Grasping. Without force feedback, the gripper fails to compensate the disturbances during robot movement, causing object slippage. With feedback control, the gripper maintains a stable grasping by actively adjusting contact force in response to disturbances.

## VI. CONCLUSION

In the paper, we introduced a small and highly sensitive FBG-based force sensor for accurate contact force measurement for gripping systems. The soft enclosed force sensor consists of a 3D-printed TPU casing with a small bump and uvula structure, a dual FBG array, and a protective tube. A series of tests were conducted to evaluate the effectiveness of the proposed force sensor, including force calibration, repeatability test, hysteresis study, force measurement comparison, and temperature calibration and compensation tests. The results demonstrated good repeatability, with a force measurement range of 4.69 N, a high sensitivity of approximately 1169.04 pm/N, a RMSE of 0.12 N, and a maximum hysteresis of 4.83%. When compared to a commercial load cell, the sensor showed a percentage error of 2.56% and an RMSE of 0.14 N. Besides, the proposed sensor validated its temperature compensation effectiveness, with an RMSE of 0.01 N over a temperature change of 11 °C. The proposed force sensor was subsequently integrated with a robotic grow-and-twine gripper to monitor interaction forces between three different objects —rubber, wood, and plastic—ranging in mass from 222 to 419 grams. The sensor monitors the gripping status between the gripper and the item. When the item begins to slip, the sensor's force reading decreases. To prevent slippage, we implement a closed-loop force control system that continuously adjusts the grip, ensuring a secure grasp throughout pick-and-place operations.

To minimize hysteresis, we may consider optimizing material selection, refining fabrication methods, or implementing software-based hysteresis compensation algorithms. While the current force sensor functions as a standalone unit, expanding it into a sensor array could broaden its applications across various fields, including manufacturing, healthcare, agriculture, logistics, and packaging.

In the future, the manual fabrication process could be transitioned to an automated system for more precise and consistent assembly. Furthermore, adjustments on both material selection and the height of the chamber can be made on a case-by-case basis to achieve the desired force measurement range. In addition to rigid objects, the force sensor is also able to interact with soft objects. Our next step is to integrate the force sensor and vision sensor with the grow-and-twine gripper and the robotic arm to facilitate multimodal sensing for automated vegetable harvesting in a vertical farm, with more sophisticated control algorithm [27].